\documentclass[conference]{ieeetran}
\IEEEoverridecommandlockouts

\usepackage{url}

\usepackage{makecell}

\usepackage{graphicx}

\usepackage{subcaption}
\usepackage[utf8]{inputenc}
\usepackage[T1]{fontenc}

\usepackage{tabularx}

\usepackage{textcomp}

\usepackage[plainpages=false,hypertexnames=false,pdfpagelabels]{hyperref} 

\begin{document}
\title{Active fall prevention: robotic vision in AAL
\thanks{This work was funded within the INCARE AAL-2017-059 project ,,Integrated Solution for Innovative Elderly Care'' by the AAL JP and co-funded by the AAL JP countries (National Centre for Research and Development, Poland under Grant AAL2/2/INCARE/2018).}}

\author{\IEEEauthorblockN{Dawid Gruszczyński}
		\IEEEauthorblockA{
			Warsaw University of Technology\\
			Institute of Control and Computation Engineering\\
			Warsaw, Poland\\
			Email: dawid.gruszczynski.stud@pw.edu.pl}
		\and
		\IEEEauthorblockN{Maciej Stefańczyk}
		\IEEEauthorblockA{
			Warsaw University of Technology\\
			Institute of Control and Computation Engineering\\
            Warsaw, Poland\\
			Email: maciej.stefanczyk@pw.edu.pl}
}
\maketitle              

\begin{abstract} 
Effective methods of preventing falls significantly improve the quality of life of the Elderly. Nowadays, people focus mainly on the proper provision of the apartment with handrails and fall detection systems once they have occurred. The article presents a~system of active detection and classification of the risk of falls in the home space using a~service robot equipped with a~vision sensor. Hazard classification allows for effective performance of tasks assigned to the robot while maintaining a~high level of user safety.

\end{abstract}
\section{Introduction}

Falls are the main cause of hospitalized injuries among the elderly \cite{eurosafe,halik2018urazy-en}. It is particularly dangerous due to osteoporosis and other diseases where the fall can be fatal. According to OECD reports \cite{oecd}, the number of people aged 65+ is growing very rapidly, and by 2030 there will be 25\% more people in the European Union than today. The number of people aged 20-64 is also crucial and remains stable, which means that the number of potential carers for older people is falling. Outlooks for 2060 are even worse, with another 25\% raise in elderly range and 13\% drop in active age range. As the population ages, the number of people living independently, even in old age, is increasing.  Preventing falls is, therefore, a~way of significantly improving their quality of life and self-confidence.  

The most common method of preventing falls in the house is to install handrails and brackets in sensitive places \cite{gillespie2012interventions}. Older people are also advised to avoid using carpets whose rolled-up edges are very easy to trip over. Similarly, the furniture used in the home is crucial. Objects with supports that protrude beyond their contours are avoided, while very stable chairs are recommended, which can be used as a~support if necessary. Unfortunately, there are also several cases that are very difficult to prevent. A~newspaper that the wind blows from a~table to the floor can cause slippage. The cable from the charger is a~danger. The number of potentially dangerous objects on the floor is also increasing because of pets (very popular among people living alone). All this makes methods of detecting this type of threat necessary to increase the level of safety. With the ever-decreasing prices of utility robots (especially cleaning robots) and their growing capabilities, it seems that equipping a~home robot with methods of detecting and identifying hazards and informing the user about them is the right direction of research. The additional possibilities can also be an advantage in the very competitive cleaning robot market.

There are solutions for solving similar problems. Kenny robot, developed in the Squirrel EU project \cite{squirrel} and similar application developed on the Toyota HRS \cite{hrs} are targeted on the toy tidying task. Robots detect the toys on the floor (only the known classes) and put them into appropriate bins using built-in manipulators. Hobbit robot~\cite{bajones2018hobbit} is designed to be used with the Elderly, and is also equipped with the manipulator. It can detect and handle the known objects, which can also be used as active fall prevention. All those systems, however, rely heavily on the manipulation skills. Although it makes the robotic platform and the whole system much more versatile, the price tag of such a system is very high, which makes it unfeasible for implementing in end-user homes. All the objects in the aforementioned systems are also treated equally, and the object detection task must be run as the main application.

In the article, we present an idea for the vision and decision system for an active fall prevention system that can be implemented on simple robots without the manipulation skills. Our system is possible to be implemented as a background task, constantly monitoring the robot's surrounding. 
If the hazardous objects are detected, the system raises the alarm, and currently executed task can be interrupted to perform appropriate action~\cite{Dudek-multitasking-romoco-2019}.

The rest of the article is structured as follows. The section~\ref{sec:state} contains an overview of existing methods of object and obstacle detection. Section~\ref{sec:system} provides the proposed system structure followed by the exemplary implementation and evaluation in section~\ref{sec:eval}. The article ends with conclusions and ideas for future extensions.

\section{State of the art} 
\label{sec:state}

\subsection{Object detection in RGB}

Fall prevention depends to a~large extent on the possibility of detecting the danger. This task is very complex due to the high randomness of
the places where potential obstacles may occur and the difficulty of their detection. In addition to this, there are many other requirements, including adaptation to dynamically changing house rooms, limitation of the need for human intervention in the work of the supervisory system, and acceptance by target users. The above requirements can only be met by using an intelligent robotic system that can properly record the elements of the environment and to distinguish between them the obstacles that pose a~risk. 

One possible approach is to use methods of object detection based on RGB environment images. Thanks to the rapid development of this field in recent years, many different models are now available to ensure high precision and speed. Even with low computing power, a~system capable of detecting a~large number of object classes in a~short time is possible. In general, when choosing the detection network, one has to decide between speed and accuracy. Two-stage solutions (Faster R-CNN or Mask R-CNN) gives better precision. The most popular single-stage detectors (SSD, YOLO) are much faster, but with worse results in terms of accuracy \cite{zhao2019object}.

Apart from the speed and precision, the important thing to take into consideration when selecting the network to use is the availability of
pre-trained models. Nowadays, for most popular applications, there are available multiple sources of those. In our scenario, where the goal
is to detect objects from everyday home environment, networks trained on the COCO (Common Objects in Context \cite{lin2014microsoft}) dataset are the way to go. 
COCO dataset contains over 120,000 pictures, with a~total of 80 different object classes labeled on them. Most are objects of the home environment, which gives an ability to detect a~broad range of potential obstacles, as well as fixed room elements such as furniture.

\subsection{Object detection in depth}

Object detectors trained on the COCO dataset are not able to find all possible interesting objects 
(e.g. there is no class for shoes, power cords, or adapters, which are the likely cause of slipping). There is 
also a~possibility to detect obstacles using point cloud data. Since this time every image pixel 
represents 3D coordinates in the camera frame, there is much more spatial data to be used.
Additionally, presuming that objects which may cause a~fall risk are located mainly on the ground, 
distinguishing them from the other environment elements can be significantly simplified. It involves only 
discarding points that belong to the plane representing the floor and segmenting out different objects 
in terms of spatial distance between them \cite{rusu20113d}. When data from the sensor is organized, 
the segmentation process can be optimized by exploiting local neighbourhood for fast region growing 
\cite{holz2013fast}, which can be further extended by adding color information to the process
\cite{stefanczyk2012multimodal}. In cluttered scenes, when we expect the objects to be man-made, it is
also a~viable option to assume their symmetry, which facilitates final segmentation \cite{ecins2016cluttered}.

\subsection{Object classification}

Objects detected in depth can be further classified to enhance the information 
for the user. This classification can be done either in RGB or based on 3D data. 
RGB classifiers don't have to cope with localization and multiple objects on the 
image, thus their accuracy is usually better than detectors. It is also a~viable
option at this point to use solutions trained on bigger datasets (like for example 
ImageNet 1000 classes). The most prominent examples of currently used architectures 
are VGG, ResNet and Inception.
The most sophisticated versions of those can reach over 80\% top-1 accuracy on challenging
datasets \cite{canziani2016analysis}, which is even higher for isolated objects 
(like in our scenario).

\section{System structure and algorithms}
\label{sec:system}

As mentioned in previous sections, RGB image or depth-based obstacle detection methods separate may not give the most accurate results. Both of them have different characteristics and are designed for slightly different tasks. However, connecting them into a~single system might result in significant error compensation. The idea is to run both methods simultaneously and compare the results to ensure detection accuracy and correctness. The main benefit of this approach is that it allows detecting obstacles that are hard to distinguish from the environment for each of those methods separately. It also does not require any additional data. A~single RGB-D sensor is enough to provide data for both processes. 

\begin{figure}[ht!]
    \centering
    \includegraphics{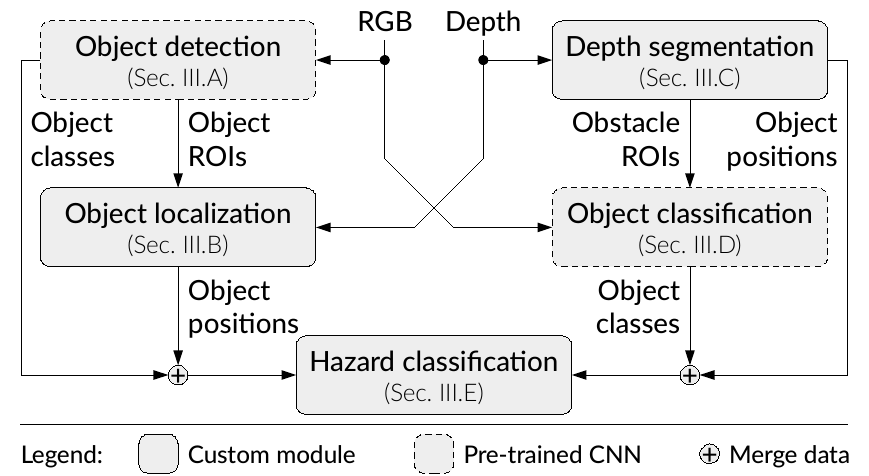}
    \caption{Complete hazard detection system structure.}
    \label{fig:system_structure}
\end{figure}

The hazard detection system, as stated before, consists of two separate processing paths: RGB image-based object detection and depth based obstacle detection. Since RGB image does not contain any information about object 3D positions, after computing detection results, the additional localization process is required. Also, the other procedure is required for determining classes of obstacle detection results. 
In the end, final results are obtained by analyzing data received from both paths. We check whether bounding boxes and classes of the same detections match, and based on that we make proper adjustments. Also, we exclude fixed environment elements such as walls and bigger furniture if it overlaps with occupied regions of the global occupation map. Finally, the hazard severity level is calculated, as some objects may pose more significant threat than others. The structure of the system is shown in Fig.~\ref{fig:system_structure} (numbers in blocks refers to the sections in the article)..

\subsection{Object detection in RGB}

The first step of the RGB-based processing path is object detection. As there are multiple available pre-trained networks available, we
decided to do a~short review of their speed and accuracy and choose the one with the best statistics. All the networks were trained on the COCO dataset so that we could compare their results based on the accuracy reported by the authors. 

\begin{table}[htb!]
    \caption{Speed and accuracy of selected pretrained object detectors.}
    \begin{tabularx}{1\columnwidth}{ |>{\centering}X|c|c|c|c|  }
        \hline
        {\textbf{Model} }     & {\textbf{Speed}} & {\textbf{mAP}}   &  {\textbf{mAP$_{50}$}}     & {\textbf{Dataset}}    \\
        \hline
        SSD\_Resnet\_50\_fpn                        &  76  &  35  & & COCO  \\
         SSDlite\_Mobilenet\_v2                      &  27  &  22  & & COCO  \\
        Faster\_RCNN\_Inception\_v2                 &  58  &  28  & & COCO  \\
        Faster\_RCNN\_Resnet50                      &  89  &  30  & & COCO  \\
        RFCN\_Resnet101                             &  92  &  30  & & COCO  \\
        \hline
        YOLOv3 ($416\times 416$)        &  28 & & 55  &  COCO  \\
        YOLOv3\_spp ($608 \times 608$)   &  50   & & 61  &  COCO  \\
        YOLOv3\_tiny ($416 \times 416$)  &  5  & & 33  &  COCO  \\
        \hline
    \end{tabularx}
    \label{tab:zestawienie}
\end{table}

The speed of the models was measured using an Nvidia GeForce GTX TITAN X graphics card. The accuracy of the models in the test set was determined using a~metric called mean Average Precision, the average precision value of the model for IoU thresholds in the ranges from 0.5 up to 0.95, while mAP$_{50}$ determines the network efficiency only for an IoU threshold of 0.5. For the YOLO networks only the latter was reported, so the results are much higher than others (Tab.~\ref{tab:zestawienie}). 

As in our scenario speed is less important than the detection accuracy, we decided to take the network with the highest reported mAP. Although 
it is impossible to directly compare mAP with mAP$_{50}$, after additional testing in the target environment, we decided to choose the spp variant of YOLOv3. We feed the color images from the head-mounted RGB-D camera as the input to the detector. As a~result list of detected objects is returned, with bounding box (in pixels) and class for each. 

\subsection{Object localization}

The next step is to determine the actual 3D location of the object. If there is no depth data available and we assume that the objects lay on the ground, we can calculate their position by intersecting the straight line from the camera center through the center of the bounding box with the ground plane. The results are very rough, but if no other data is available, then it is the easiest solution. If the depth data is available from the sensor, the localization can be more accurate. At first, the object must be segmented from the background, and after that, the created mask could be used to calculate the position. It is possible to use the detector returning the masks (like Mask R-CNN), but those usually have lower detection accuracy. Alternatively, the masks can be created by segmenting the detected obstacles directly in the depth image. This approach is less accurate, as some parts of the background leak to the object segment, but is more versatile, as every object detector returns bounding boxes, and only some of them return masks. 

In our work, we implemented three different methods of depth image segmentation: segmentation using the histogram,  using the K-means clustering algorithm, and using double thresholding based on the average distance from the camera (removing background and foreground objects, if any).

Segmentation using a~histogram is based on the assumption that the object occupies most of the analyzed area, or is its most coherent region. In most cases, this is fulfilled because the frames from the detector are usually well aligned with the detected objects.
Depth histogram is divided into $k$ bins of equal width, and the one with the highest number of assigned pixels is selected as the final distance. 

Segmentation using the K-means clustering algorithm is also based on the assumption that the object is the dominant element of the analyzed area. As before, this method determines several pixel depth value clusters, but this time the range of each cluster is dynamically selected using the centroids algorithm and $k$ clusters.  The final distance is calculated as the average distance in the largest cluster.

\begin{figure}[ht!]
    \centering
    \includegraphics{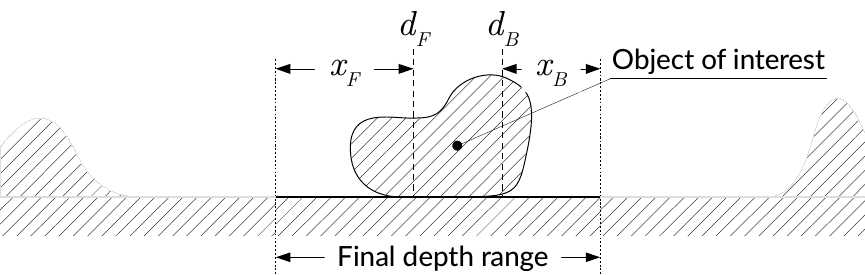}
    \caption{Double thresholding for object segmentation}
    \label{fig:thr}
\end{figure}

Segmentation using double thresholding based on the average distance from the camera is different from other methods. It uses the assumption that an object may have a~background behind it, or may be obscured by small objects in the foreground. Thus, the area occupied by the object does not have to be dominant in the examined area. At first, the average distance $d_B$ is calculated for all the points inside the ROI. After that, the background is removed by rejecting all points with a~distance greater than $d_B + x_B$, where $x_B$ is a~tunable parameter. The foreground is removed similarly. After removing the background, the average distance $d_F$ is calculated, and all points closer than $d_F - x_F$ are rejected. What remains is taken as the object, and the final distance is the average of its pixels (Fig.~\ref{fig:thr}).

The distance from the camera to the detected object, determined using segmentation methods described above, is the basis for calculating the final position of the object in the environment. For this purpose first the object position in the camera frame is calculated from using the
camera intrinsics: 
$$X = (x - c_x)\cdot\frac{Z}{f_x}\quad Y = (y - c_y)\cdot\frac{Z}{f_y}$$
where $Z$ is the average distance to the object and $x, y$ is the center of the objects bounding box. After that the position is transformed to the global coordinate frame using the robot position and head orientation from the control system.

\subsection{Depth based obstacle detection}

Depth-based path of the presented system starts from the obstacle detection. The initial assumption is that the robot is looking at the floor, and the possible sources of Elderly fall are lying there. The obstacle detection process relies on the floor plane segmentation as the first step.
This process is done using the RanSaC algorithm, which fits a~plane model to the input pointcloud (created from the depth image). The plane found is considered to be a~floor on which obstacles are located. Each point belonging to it is then removed from the point cloud.

After removing the plane from the point cloud, usually only few groups of points remain. They represent potential obstacles, and to distinguish between them, another segmentation process is required. This time the aim is to find groups of points which belong to the same regions. The way to achieve that is to apply the Region Growing Segmentation algorithm. It combines into clusters points from consistent surfaces, and thus depth readings from the same objects. Finally, the segmented points are used to create the masks for the objects by projecting them back to the image space. Object positions are determined by calculating the centroids of all the points for every object.

\subsection{Obstacle classification}

The main disadvantage of the obstacle detection process using point cloud data is the lack of information about classes of detected objects. As we have created the masks for the objects in image space, it is possible now to cut interesting parts from the RGB image and pass them to the object classifier. For this purpose, similar to the object detection, we also utilized the pre-trained networks available in the \textit{Tensorflow} repositories, this time trained on the ImageNet database. Based on the accuracy comparison in \cite{canziani2016analysis}, we decided to use the Inception-v3 model due to its high \mbox{top-1} accuracy and moderate speed. The last decision to be made here was how to cut the objects from the image. When we used the masks to create the bounding box and cut tight images with only the object and almost no surrounding, the results had low accuracy. The proper way, for this network, was to cut the object with a~lot of surrounding area (we achieved the best results with ROIs 3 times bigger than the object itself). 

\subsection{Hazard classification}

Up to this point, all the information (either from RGB or depth paths) is in the form of labeled objects and their positions. Without further processing, this is only slightly more useful than simple obstacle detection. The critical part of the whole system is hazard classification. From the detected objects, we have to select the obstacles that pose the highest risk and ignore the safe ones. At this point, the rule-based system (based on simple ontology and semantic map of the environment) was created. At first, all the results from color and depth paths are merged to remove duplicates. After that the objects classified as furniture elements are checked whether they are standing near the walls. If some furniture is found near the center of the room, it is treated as a~potential threat. All the objects that are placed higher than the floor are also discarded, as they are possibly detected on the table. The remaining objects are divided into animals and other things. As animals are (usually) moving from place to place, they are marked as moderate risk. All other things are treated as high risk and reported to the user.

\section{Evaluation}
\label{sec:eval}

In our tests, we used the TIAGo robot as a~source of data. It is a~mobile service robot with a~differential drive, designed to work in a~home environment and to serve people. It is equipped with an Orbec Astra RGB-D sensor placed on an articulated head, which allows us to collect necessary data for a~fall prevention task easily. Our system was implemented using the ROS framework, which is also the official way to control the TIAGo robot. Presented solution is implemented as a module in the bigger control system, which is based on the structure presented in~\cite{zielinski2017variable-twiki}.

\begin{figure}[ht!]
    \centering
    \includegraphics[width=\columnwidth,trim=0 0 0 2.4cm,clip]{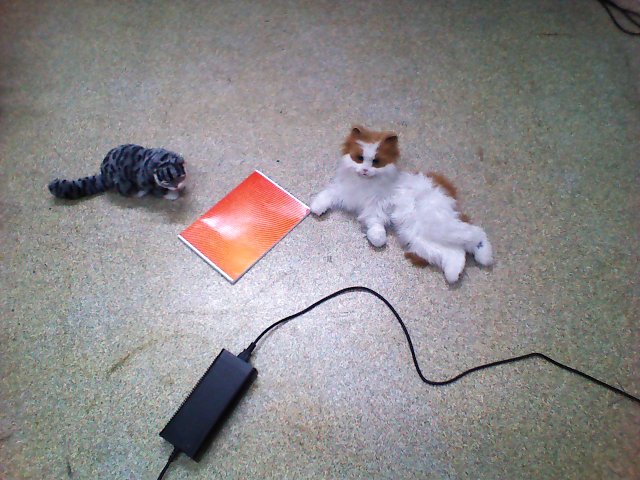}
    \caption{Arranged test scene}
    \label{fig:objectsa}
\end{figure}

PAL Robotics (the maker of the robot), apart from the robot control system, also provides the full-featured simulator based on the Gazebo. We used it for the first experiments to prove the correctness of our approach. After that the simple test scene was arranged to cover all the expected cases of our system (Fig.~\ref{fig:objectsa}). In particular, we have two animals on the scene (both should be detected by the RGB and depth paths), thin book (to thin to be detected in depth) and power adapter (that kind of object doesn't exist in COCO dataset, so should be ignored by the RGB detector).

\subsection{RGB image based detection path}

\begin{figure}[ht!]
    \centering
    \includegraphics[width=\columnwidth,trim=0 0 0 2.4cm,clip]{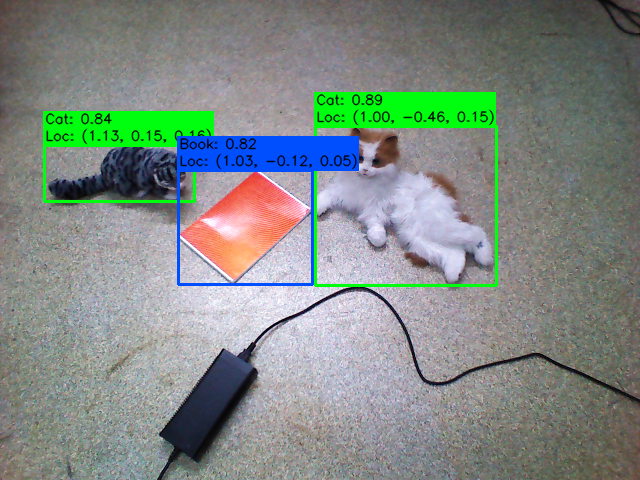}
    \caption{RGB image based detection path results}
    \label{fig:objectsb}
\end{figure}

At first the RGB object detection path was tested. As we expected, both cats and the book were detected with appropriate classes assigned, and the power adapter was ignored (Fig.~\ref{fig:objectsb}). In this scene power adapter with dangling cord is presumably the biggest risk.

\begin{figure}[htb!]
~~~~\includegraphics[width=0.2\textwidth]{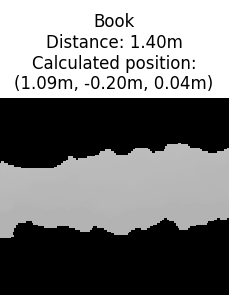}\hfill%
\includegraphics[width=0.2\textwidth]{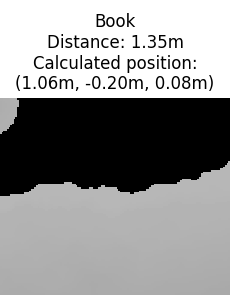}\\
a) \includegraphics[width=0.2\textwidth]{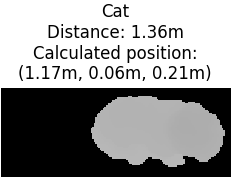}\hfill%
b) \includegraphics[width=0.2\textwidth]{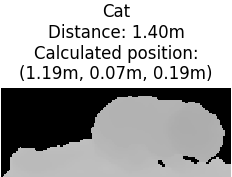}\\
\caption{Segmentation results for: a) histogram, b) double thresholding}
\label{fig:segmentation_results}
\end{figure}

Presented objects' localizations are based on the segmentation results. Bounding boxes of each detected object are projected onto the depth image and then segmented by the three separate methods described in section \ref{sec:system}. Sample segmentation results for the smaller cat (easy case) and book (hard case) are presented in Fig.~\ref{fig:segmentation_results}. Despite leaving part of the ground in the front, depth values are consistent, meaning that it has a~negligible impact on the final calculation of the distance from the camera. The only problematic object, in this case, was the book. However, even if the distance is mainly calculated based on the nearby floor readings, it still does not introduce a~significant error.

\subsection{Depth based detection path}

\begin{figure}[h]
    \centering
    \includegraphics[width=\columnwidth,trim=0 0 0 2.4cm,clip]{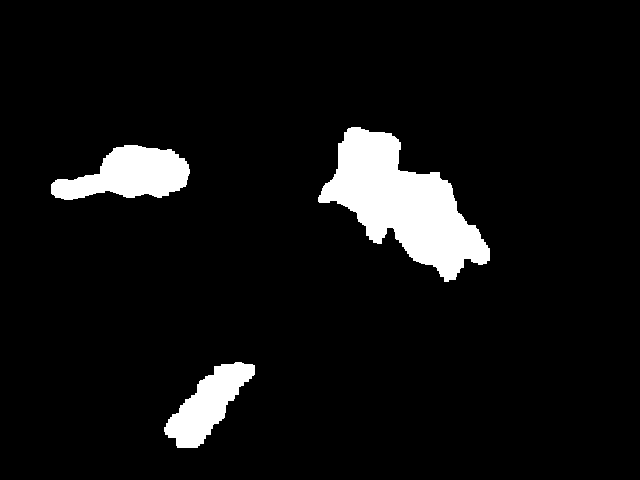}
    \caption{Depth based detection path masks}
    \label{fig:deptha}
\end{figure}

\begin{figure}[h]
    \centering\includegraphics[width=\columnwidth,trim=0 0 0 2.4cm,clip]{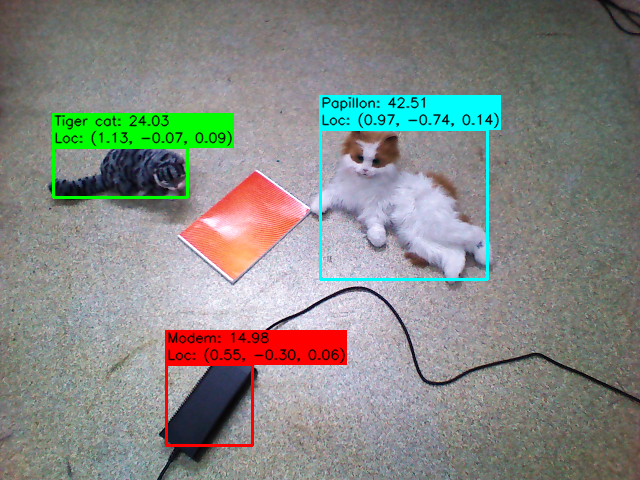}
    \caption{Depth based detection classes}
    \label{fig:depthb}
\end{figure}

For the same scene as before, the depth-based obstacle detection was applied. Detected objects masks are presented in Fig.~\ref{fig:deptha}.
This time the book is missing in the results (as expected), but other objects were segmented properly. Object classification results are presented in Fig.~\ref{fig:depthb}. As in the ImageNet classes there are different labels for many dog and cat breeds, this time the animals were classified as those. Although the bigger cat is classified as \textit{papillon} (dog breed, very similar to the presented cat), the general \textit{animal} category is enough for our purpose. This time the power supply has been successfully detected and the class assigned to it was quite accurate (since modem has a~similar shape).

\subsection{System outcome}
\begin{figure}[htb!]
    \centering
    \includegraphics[width=\columnwidth,trim=0 0 0 2.4cm,clip]{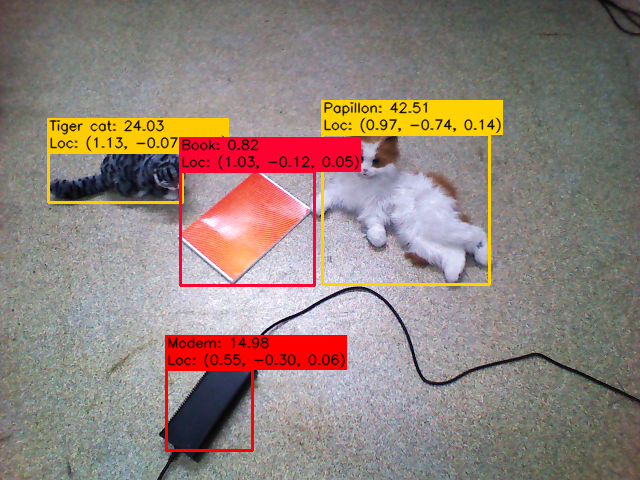}
    \caption{Detected hazards with their classification for moderate (cats) and high risk of fall (power adapter and book)}
    \label{fig:hazards}
    \end{figure}

The results from both paths are duplicated for both cats -- those could be merged at the last step. As expected, the objects not existing in the COCO dataset are ignored in RGB, but the depth part can properly recognize them. As the last step, we assign the severity level for each object. In the sample scene, the cats are assigned with moderate severity and power adapter and book with the high (Fig.~\ref{fig:hazards}). 

\section{Summary}

The article presented the system of obstacle detection in application for active fall prevention. Obstacles are detected in parallel in color and depth data, which makes the final results better than using any one of them alone. The important part of the system is fall hazard severity classification. This can be used as a~measure of whether the robot can continue its current task and inform the user later, or the current task must be suspended and the user must be notified as soon as possible. 
Obstacle classification can also be used by other control modules, for example, path planning. In potential field approaches~\cite{mmar_seredynski_graph-2016}, the repulsive force can be smaller for static obstacles and higher for animals (as those can move just before the robot). Also, the global planners can use this information to take into account the possibility of driving through the door blocked by some objects. In that case, lack of direct manipulation skills of the robot can be mitigated by utilizing pushing skills~\cite{krivic2018online} to move the objects out of the way.
System can be easily extended with other types of hazards, like detecting opened door or light left switched on (or off in the evening)~\cite{mmar_dudek_distributed-2016}. Prepared algorithms are going to be used as the part of the robotic platform in the Incare project~\cite{incare-www-wut}.

The acceptance level of such a~system is an open question. From the analysis of fall detection systems, it can be concluded that people are quite open and willing to use wearable fall detection devices \cite{kolakowski2020localization}, while the acceptance of continuous living room monitoring is quite low  \cite{sitar2018smart}. Equipping a~home with an additional sensor is more acceptable. It can be assumed that a~mobile fall-risk detection sensor that scans only part of the floor and is activated on demand will be easy to accept.

\bibliographystyle{IEEEtran}
\bibliography{ref}

\begin{thebibliography}{10}
\providecommand{\url}[1]{#1}
\csname url@samestyle\endcsname
\providecommand{\newblock}{\relax}
\providecommand{\bibinfo}[2]{#2}
\providecommand{\BIBentrySTDinterwordspacing}{\spaceskip=0pt\relax}
\providecommand{\BIBentryALTinterwordstretchfactor}{4}
\providecommand{\BIBentryALTinterwordspacing}{\spaceskip=\fontdimen2\font plus
\BIBentryALTinterwordstretchfactor\fontdimen3\font minus
  \fontdimen4\font\relax}
\providecommand{\BIBforeignlanguage}[2]{{%
\expandafter\ifx\csname l@#1\endcsname\relax
\typeout{** WARNING: IEEEtran.bst: No hyphenation pattern has been}%
\typeout{** loaded for the language `#1'. Using the pattern for}%
\typeout{** the default language instead.}%
\else
\language=\csname l@#1\endcsname
\fi
#2}}
\providecommand{\BIBdecl}{\relax}
\BIBdecl

\bibitem{eurosafe}
EuroSafe, ``{Injuries in the European Union. Summary of injuries statistics for
  the years 2012-2014}.''

\bibitem{halik2018urazy-en}
R.~Halik, J.~Stokwiszewski, B.~Wojtyniak, and W.~Seroka, \emph{Injuries to
  people over 60 years old in {P}oland}.\hskip 1em plus 0.5em minus 0.4em\relax
  National Institute of Public Health - National Institute of Hygiene, 2018,
  (report in Polish).

\bibitem{oecd}
OECD, ``Population projections,'' in \emph{Demography and population}.\hskip
  1em plus 0.5em minus 0.4em\relax OECD Publishing, 2020,
  \url{https://stats.oecd.org/Index.aspx?DataSetCode=POP_PROJ#}, (accessed 5
  march 2020).

\bibitem{gillespie2012interventions}
L.~D. Gillespie, M.~C. Robertson, W.~J. Gillespie, C.~Sherrington, S.~Gates,
  L.~M. Clemson, and S.~E. Lamb, ``Interventions for preventing falls in older
  people living in the community,'' \emph{Cochrane database of systematic
  reviews}, no.~9, 2012.

\bibitem{squirrel}
\BIBentryALTinterwordspacing
{EU FP7 grant}, ``Squirrel: Clearing clutter bit by bit,'' 2014-2018. [Online].
  Available: \url{http://www.squirrel-project.eu}
\BIBentrySTDinterwordspacing

\bibitem{hrs}
\BIBentryALTinterwordspacing
{Preferred Networks Inc.}, ``Autonomous tidying-up robot system,'' CEATEC JAPAN
  2018. [Online]. Available:
  \url{https://projects.preferred.jp/tidying-up-robot/en/}
\BIBentrySTDinterwordspacing

\bibitem{bajones2018hobbit}
M.~Bajones, D.~Fischinger, A.~Weiss, D.~Wolf, M.~Vincze, P.~de~la Puente,
  T.~K{\"o}rtner, M.~Weninger, K.~Papoutsakis, D.~Michel \emph{et~al.},
  ``Hobbit: Providing fall detection and prevention for the elderly in the real
  world,'' \emph{Journal of Robotics}, vol. 2018, 2018.

\bibitem{Dudek-multitasking-romoco-2019}
W.~Dudek, M.~Węgierek, J.~Karwowski, W.~Szynkiewicz, and T.~Winiarski, ``Task
  harmonisation for a single--task robot controller,'' in \emph{12th
  International Workshop on Robot Motion and Control (RoMoCo)}.\hskip 1em plus
  0.5em minus 0.4em\relax IEEE, 2019, pp. 86--91.

\bibitem{zhao2019object}
Z.-Q. Zhao, P.~Zheng, S.-t. Xu, and X.~Wu, ``Object detection with deep
  learning: A~review,'' \emph{IEEE transactions on neural networks and learning
  systems}, vol.~30, no.~11, pp. 3212--3232, 2019.

\bibitem{lin2014microsoft}
T.-Y. Lin, M.~Maire, S.~Belongie, J.~Hays, P.~Perona, D.~Ramanan,
  P.~Doll{\'a}r, and C.~L. Zitnick, ``Microsoft coco: Common objects in
  context,'' in \emph{European conference on computer vision}.\hskip 1em plus
  0.5em minus 0.4em\relax Springer, 2014, pp. 740--755.

\bibitem{rusu20113d}
R.~B. Rusu and S.~Cousins, ``3d is here: Point cloud library (pcl),'' in
  \emph{2011 IEEE international conference on robotics and automation}.\hskip
  1em plus 0.5em minus 0.4em\relax IEEE, 2011, pp. 1--4.

\bibitem{holz2013fast}
D.~Holz and S.~Behnke, ``Fast range image segmentation and smoothing using
  approximate surface reconstruction and region growing,'' in \emph{Intelligent
  autonomous systems 12}.\hskip 1em plus 0.5em minus 0.4em\relax Springer,
  2013, pp. 61--73.

\bibitem{stefanczyk2012multimodal}
M.~Stefa{\'n}czyk and W.~Kasprzak, ``Multimodal segmentation of dense depth
  maps and associated color information,'' in \emph{International Conference on
  Computer Vision and Graphics}.\hskip 1em plus 0.5em minus 0.4em\relax
  Springer, 2012, pp. 626--632.

\bibitem{ecins2016cluttered}
A.~Ecins, C.~Ferm{\"u}ller, and Y.~Aloimonos, ``Cluttered scene segmentation
  using the symmetry constraint,'' in \emph{2016 IEEE International Conference
  on Robotics and Automation (ICRA)}.\hskip 1em plus 0.5em minus 0.4em\relax
  IEEE, 2016, pp. 2271--2278.

\bibitem{canziani2016analysis}
A.~Canziani, A.~Paszke, and E.~Culurciello, ``An analysis of deep neural
  network models for practical applications,'' \emph{arXiv preprint
  arXiv:1605.07678}, 2016.

\bibitem{zielinski2017variable-twiki}
C.~Zieliński, M.~Stefańczyk, T.~Kornuta, M.~Figat, W.~Dudek, W.~Szynkiewicz,
  W.~Kasprzak, J.~Figat, M.~Szlenk, T.~Winiarski, K.~Banachowicz,
  T.~Zielińska, E.~G. Tsardoulias, A.~L. Symeonidis, F.~E. Psomopoulos, A.~M.
  Kintsakis, P.~A. Mitkas, A.~Thallas, S.~E. Reppou, G.~T. Karagiannis,
  K.~Panayiotou, V.~Prunet, M.~Serrano, J.-P. Merlet, S.~Arampatzis, A.~Giokas,
  L.~Penteridis, I.~Trochidis, D.~Daney, and M.~Iturburu, ``Variable structure
  robot control systems: The rapp approach,'' \emph{Robotics and Autonomous
  Systems}, vol.~94, pp. 226 -- 244, 2017.

\bibitem{mmar_seredynski_graph-2016}
D.~Seredyński, K.~Banachowicz, and T.~Winiarski, ``{Graph–based potential
  field for the end–effector control within the torque–based task
  hierarchy},'' in \emph{21th IEEE International Conference on Methods and
  Models in Automation and Robotics, MMAR'2016}.\hskip 1em plus 0.5em minus
  0.4em\relax IEEE, 2016, pp. 645--650.

\bibitem{krivic2018online}
S.~Krivic and J.~Piater, ``Online adaptation of robot pushing control to object
  properties,'' in \emph{2018 IEEE/RSJ International Conference on Intelligent
  Robots and Systems (IROS)}.\hskip 1em plus 0.5em minus 0.4em\relax IEEE,
  2018, pp. 4614--4621.

\bibitem{mmar_dudek_distributed-2016}
W.~Dudek, K.~Banachowicz, W.~Szynkiewicz, and T.~Winiarski, ``{Distributed NAO
  robot navigation system in the hazard detection application},'' in \emph{21th
  IEEE International Conference on Methods and Models in Automation and
  Robotics, MMAR'2016}.\hskip 1em plus 0.5em minus 0.4em\relax IEEE, 2016, pp.
  942--947.

\bibitem{incare-www-wut}
\BIBentryALTinterwordspacing
``{INCARE project page at WUT},'' Mar. 2020. [Online]. Available:
  \url{https://www.robotyka.ia.pw.edu.pl/projects/incare/}
\BIBentrySTDinterwordspacing

\bibitem{kolakowski2020localization}
J.~Kolakowski, V.~Djaja-Josko, M.~Kolakowski, and J.~Cichocki, ``Localization
  system supporting people with cognitive impairment and their caregivers,''
  \emph{International Journal of Electronics and Telecommunications}, vol.~66,
  no.~1, pp. 125--131, 2020.

\bibitem{sitar2018smart}
A.-V. Sitar-Taut, D.-A. Sitar-Taut, O.~Cramariuc, V.~Negrean, D.~Sampelean,
  L.~Rusu, O.~Orasan, A.~Fodor, G.~Dogaru, and A.~Cozma, ``Smart homes for
  older people involved in rehabilitation activities-reality or dream,
  acceptance or rejection,'' \emph{Balneo Res J}, vol.~9, no.~3, pp. 291--8,
  2018.

\end{thebibliography}

\end{document}